\documentclass[10pt,twocolumn,letterpaper]{article}

\usepackage[final,algorithms]{wacv}      

\usepackage{lineno}
\usepackage{graphicx}
\usepackage{amsmath}
\usepackage{amssymb}
\usepackage{booktabs}
\usepackage{comment}
\usepackage{svg}
\usepackage{tabularx}
\usepackage{multicol}
\usepackage{multirow}
\usepackage{bbm}
\usepackage{algpseudocode}
\usepackage[pagebackref,breaklinks,colorlinks]{hyperref}
\usepackage[capitalize]{cleveref}
\usepackage[ruled,vlined]{algorithm2e} 
\usepackage{xcolor}
\usepackage{fancyhdr}
\usepackage{booktabs}
\usepackage{listings}

\definecolor{codegreen}{rgb}{0,0.6,0}
\definecolor{codegray}{rgb}{0.5,0.5,0.5}
\definecolor{codepurple}{rgb}{0.58,0,0.82}
\definecolor{backcolour}{rgb}{1, 1, 1}

\lstdefinestyle{mystyle}{
    backgroundcolor=\color{backcolour},
    commentstyle=\color{codegreen},
    keywordstyle=\color{magenta},
    stringstyle=\color{codepurple},
    basicstyle=\ttfamily\footnotesize,
    breakatwhitespace=false,         
    breaklines=true,                 
    captionpos=b,                    
    keepspaces=true,                 
    numbersep=5pt,                  
    showspaces=false,                
    showstringspaces=false,
    showtabs=false,                  
    tabsize=2
}
\crefname{section}{Sec.}{Secs.}
\Crefname{section}{Section}{Sections}
\Crefname{table}{Table}{Tables}
\crefname{table}{Tab.}{Tabs.}

\def\confName{WACV}
\def\confYear{2026}

\title{RobustFormer: Noise-Robust Pre-training for Images and Videos}

\author{
    Ashish Bastola$^{1,}$\thanks{Equal contribution} , Nishant Luitel$^{2,}$\footnotemark[1] , Hao Wang$^{1}$, Danda Pani Paudel$^{2}$,
    Roshni Poudel$^{2}$,  Abolfazl Razi$^{1}$ \\
    $^{1}$Clemson University \quad $^{2}$NAAMII \\
    {\tt\small \{abastol,hao9,arazi\}@clemson.edu, \{nishant.luitel,danda.paudel,roshani.poudel\}@naamii.org.np}
}

\begin{document}
\maketitle

\begin{abstract}
While deep learning-based models like transformers, have revolutionized time-series and vision tasks, they remain highly susceptible to noise and often overfit on noisy patterns rather than robust features. This issue is exacerbated in vision transformers, which rely on pixel-level details that can easily be corrupt. To address this, we leverage the discrete wavelet transform (DWT) for its ability to decompose into multi-resolution layers, isolating noise primarily in the high frequency domain while preserving essential low-frequency information for resilient feature learning. Conventional DWT-based methods, however, struggle with computational inefficiencies due to the requirement for a subsequent inverse discrete wavelet transform (IDWT) step. In this work, we introduce RobustFormer, a novel framework that enables noise-robust masked autoencoder (MAE) pre-training for both images and videos by using DWT for efficient downsampling, eliminating the need for expensive IDWT reconstruction and simplifying the attention mechanism to focus on noise-resilient multi-scale representations. To our knowledge, RobustFormer is the first DWT-based method fully compatible with video inputs and MAE-style pre-training. Extensive experiments on noisy image and video datasets demonstrate that our approach achieves up to 8\% increase in Top-1 classification accuracy under severe noise conditions in Imagenet-C and up to 2.7\% in Imagenet-P standard benchmarks compared to the baseline and up to 13\% higher Top-1 accuracy on UCF-101 under severe custom noise perturbations while maintaining similar accuracy scores for clean datasets. We also observe the reduction of computation complexity by up to 4.4\% through IDWT removal compared to VideoMAE baseline without any performance drop.
\end{abstract}

\section{Introduction}
\label{sec:intro}
 \begin{figure}[htbp]
     \centering
     \includegraphics[width=0.9\linewidth]{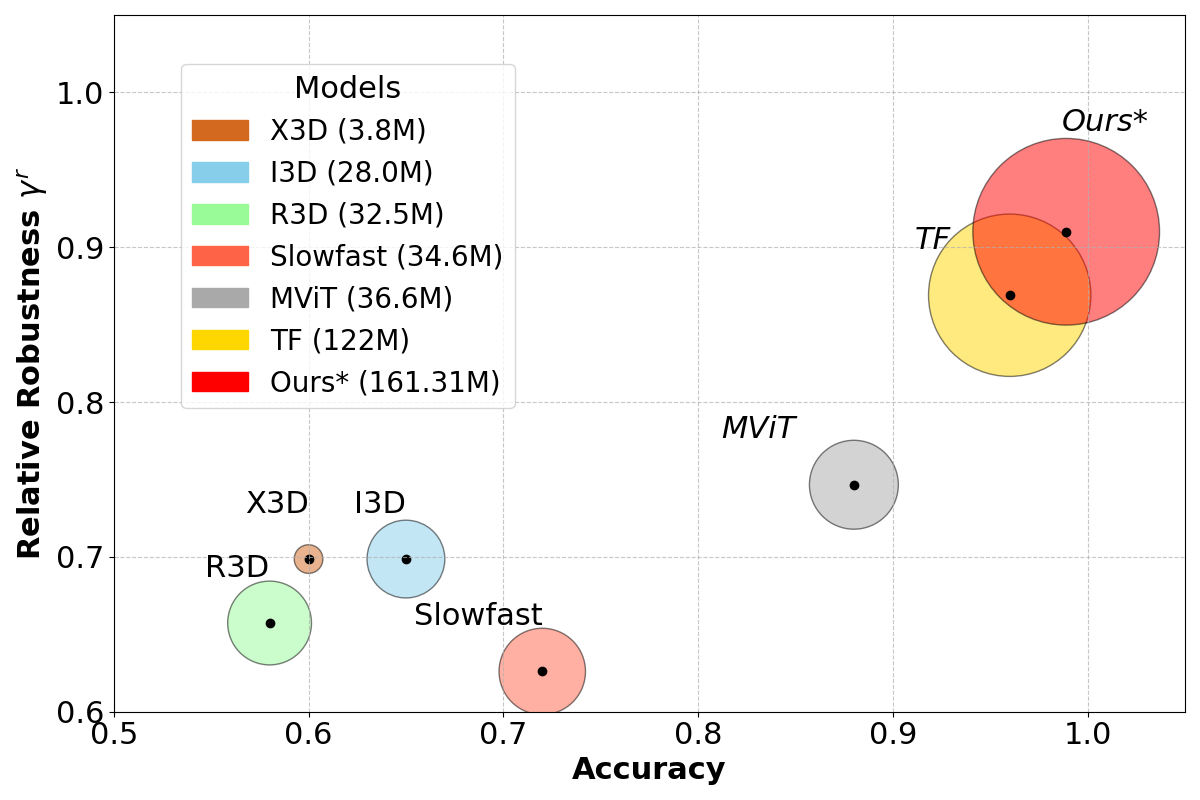}
    \caption{Accuracy vs. relative robustness (performance on corrupted vs. clean data) of action recognition models on UCF-101.}
     \label{fig:teaser}
 \end{figure}

Addressing robustness in deep learning models for images and video data is crucial for practical applications, including surveillance, autonomous driving, and multimedia content analysis \cite{liu2022towards}. Images can suffer from distortions such as blurring, noise, and artifacts, while video data experience similar issues, compounded by the additional variability introduced over time \cite{pan2024svastin}. 
Deep learning models, especially those trained on clean, high-quality data, are highly sensitive to these disruptions because they rely heavily on precise patterns within the data to make accurate predictions \cite{tong2022videomae}. Even minor noise or inconsistencies in an image can cause the model to misinterpret important features, as these models are not inherently robust to unexpected distortions \cite{schiappa2023large}. With video data, the challenge becomes even greater due to the temporal nature of the information, where each frame builds upon previous ones to create a coherent sequence \cite{wu2024waveformer}. 
For instance, temporal inconsistencies such as random noise across frames, motion blur, flickering shadows, and heat-induced turbulence can lead to abrupt changes between frames, confusing the model as it attempts to track and interpret motion or recognize actions \cite{feichtenhofer2022masked,wu2024waveformer}. 
In tasks like autonomous driving, these inconsistencies are highly prevalent and pose significant risks, as the model may fail to accurately detect objects, assess distances, or predict movement patterns \cite{razi2023deep}.

A primary reason for these issues lies in the pixel-based design of most deep-learning models. Although RGB pixel representations are the standard in vision applications because they capture detailed color and spatial information, they are computationally demanding and highly sensitive to noise and domain shifts. This sensitivity arises since RGB data represent exact pixel-level details, so even slight distortions can disrupt deep learning models that depend on these patterns for accuracy. Additionally, RGB representations process both essential and irrelevant features equally, limiting their efficiency. Inspired by image compression techniques like JPEG, frequency domain transformations such as Discrete Cosine Transform (DCT) provide a solution by isolating low-frequency components (representing significant information) from high-frequency ones (less relevant details) \cite{li2023discrete, park2023rgb}. This separation reduces memory demands and increases noise resistance without compromising performance \cite{eskicioglu1995image, gao2021neural, li2023discrete}.

While pure frequency domain transformations offer these benefits, they also lead to a loss of spatial or temporal locality due to the change in basis, a consequence of the Heisenberg uncertainty principle. In contrast, Discrete Wavelet Transform (DWT) provides a more balanced approach, decomposing the signal into both high- and low-frequency components across multiple scales. This multi-resolution representation allows DWT to retain spatial and temporal information, preserving important features and contextual details while still isolating high-frequency information. Image-based methods that leverage DWT have shown promise in improving robustness against spatial noise by allowing models to focus on frequency-based features that are less affected by disturbances \cite{Li_2020_CVPR}. Applying these principles to images and videoscan address both spatial and temporal (for videos) corruptions, as DWT can help models focus on low-frequency information, improving robustness to noise and other distortions across frames.

In this paper, we employ a masked autoencoding approach \cite{NEURIPS2022_416f9cb3, Wang_2023_CVPR}, additionally, incorporating DWT for pre-training and fine-tuning on the uncorrupted datasets, followed by evaluation on their corrupted counterparts. We experiment on widely used datasets for both images and videos, where the generated corruptions include both spatial and temporal types reflecting real-world scenarios \cite{yi2021benchmarking}. 

In summary, our contributions are as follows:\\
    \hspace*{1.5em}$\bullet$ \hspace{0.3em} We present a novel Discrete Wavelet Transform (DWT) based masked autoencoder architecture that is robust to spatial and temporal corruptions in both video and image data. To our knowledge, this is the first work to implement DWT in a masked autoencoder setting.\\
    \hspace*{1.5em}$\bullet$ \hspace{0.3em} We perform a comprehensive evaluation of several real-world noise types with varying severity levels in large-scale benchmark datasets.\\
    \hspace*{1.5em}$\bullet$ \hspace{0.3em} We demonstrate that our method performs equally well and in many cases better than the commonly used IDWT variant, which requires comparatively more compute resource and makes architectures overly complicated.

\section{Related Works}
\label{sec:related_works}

\subsection{Masked Video Training}

BERT introduced the concept of masked language modeling (MLM), a novel pre-training objective for natural language understanding, which led to significant advances in NLP \cite{kenton2019bert} and further inspiring extensions including RoBERTa \cite{Liu2019RoBERTaAR}, ALBERT \cite{Lan2019ALBERTAL}, and ELECTRA \cite{clark2020electra}, extended this masked training approach. Also inspired by BERT's masked token prediction, the concept of masked image modeling (MIM) emerged for computer vision tasks. iGPT \cite{Chen2020GenerativePF} first adapted the Transformer model for image data by treating images as pixel sequences and using a similar masked prediction task. BEiT \cite{bao2022beitbertpretrainingimage} extended this concept by proposing a discrete variational autoencoder (dVAE) to tokenize image patches, predicting masked tokens in a BERT-style pre-training. These approaches demonstrated that MIM could effectively learn transferable representations for various downstream tasks. Masked Autoencoders (MAE) \cite{He_2022_CVPR} refined MIM by reconstructing masked regions in images, achieving state-of-the-art results on multiple benchmarks using Vision Transformers (ViTs) and large-scale unlabeled data.

Building on masked training in language and image domains, Masked Video Modeling (MVM) has emerged to capture both temporal and spatial representations from video data. VideoMAE \cite{NEURIPS2022_416f9cb3} extends MAE to videos by applying random masking across both spatial and temporal dimensions, enabling the model to learn robust spatio-temporal representations, as shown in Figure \ref{fig:compare} (a). VideoMAE achieves state-of-the-art performance on numerous video classification tasks, highlighting the effectiveness of masked training for video data. Similarly, BEVT (BEiT for Video) \cite{Wang_2022_CVPR} extends BEiT's tokenization and masked prediction strategy to video patches, achieving strong results in action recognition tasks. 

\subsection{Discrete Wavelet transform(DWT)}

Wavelets are essential in time-frequency analysis and signal processing tasks, such as anti-aliasing and detail restoration, through Discrete Wavelet Transform (DWT) and its inverse (IDWT) \cite{mallat1989theory}. Early studies combined wavelets with shallow neural networks for function approximation and classification, optimizing wavelet parameters within the network \cite{szu1992neural}. Recently, deeper networks have adopted wavelets for image classification, although these integrations can be computationally intensive \cite{de2020multi}. The Multilevel Wavelet CNN (MWCNN) \cite{Liu_2018_CVPR_Workshops} applies Wavelet Packet Transform (WPT) for image restoration, handling both low- and high-frequency components. Similarly, the Convolutional-Wavelet Neural Network (CWNN) \cite{DUAN2017255} utilizes dual-tree complex wavelet transform to reduce noise in SAR images while preserving crucial features, though within a simplified two-layer structure. Wavelet Pooling \cite{williams2018wavelet} employs a two-level DWT for pooling and combines DWT/IDWT for backpropagation, deviating from conventional gradient methods.

A key advantage of wavelets is their ability to provide sparse representations, which enables more efficient processing and faster model training \cite{williams2018wavelet,li2020wavelet,wu2024waveformer}. DWT-based training reduces redundancy, making it well-suited for real-time video classification tasks where small shifts in data can significantly affect model output. Unlike other approaches, our method avoids reconstruction after the wavelet transform, thus saving considerable computational resources, especially when handling images and videos. This streamlined process enhances efficiency while maintaining robust performance in noisy environments.

\subsection{Noise Robustness}
Robustness in video classification has been an ongoing research focus, initially centered on adapting image robustness techniques to video data. For instance, \cite{hendrycks2018benchmarking} introduced benchmarks for evaluating image classifiers’ robustness against corruptions, a framework later adapted to videos by \cite{hirata2021making} to assess model performance under various perturbations. Addressing the challenge of temporal consistency in video data, models such as the two-stream network \cite{NIPS2014_00ec53c4} and 3D convolutions \cite{8099985} were developed, with data augmentation methods like temporal jittering and frame dropping \cite{Shorten2019ASO} further enhancing robustness. 

\begin{figure}[htbp]
    \centering
    \includegraphics[width=\linewidth]{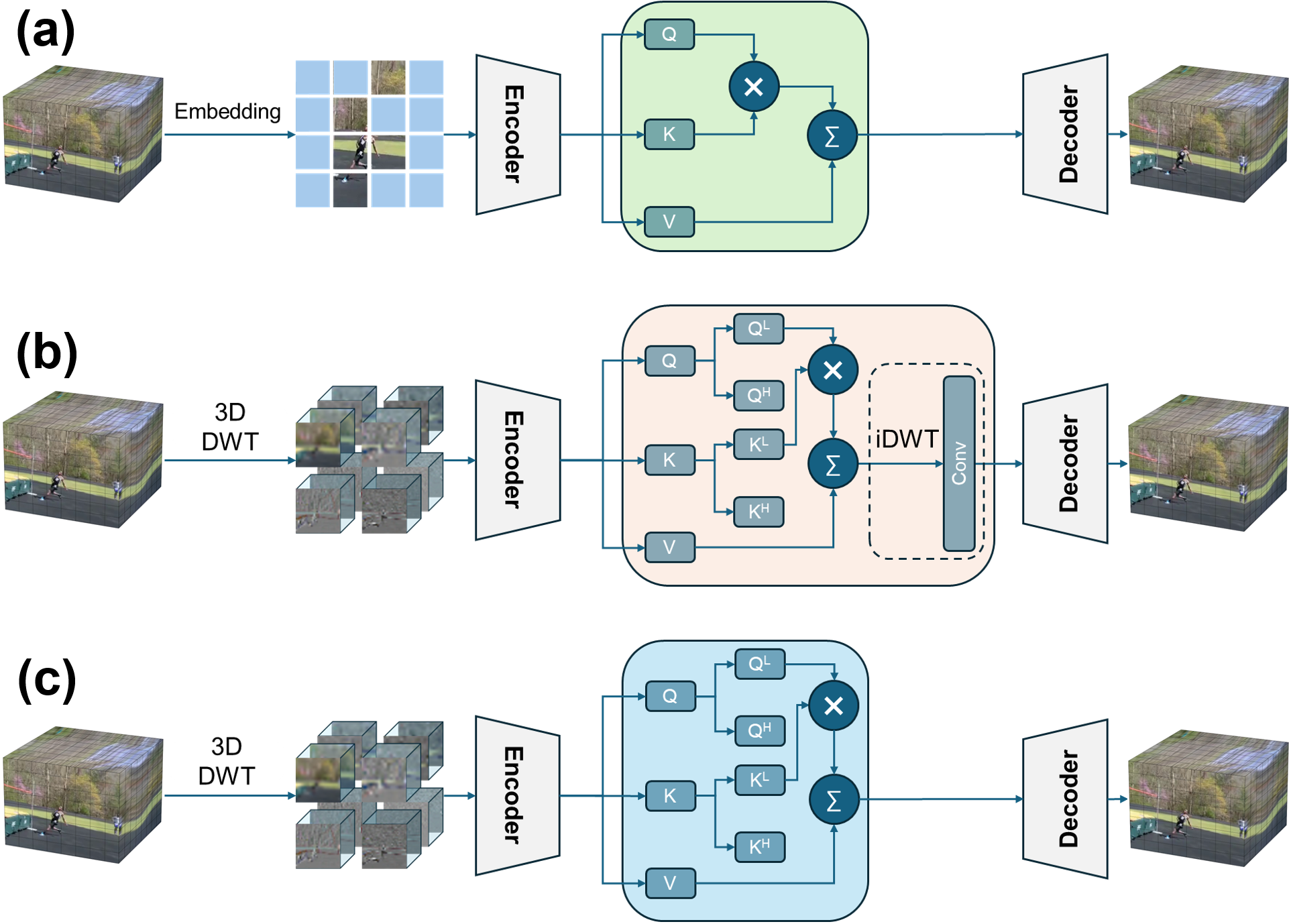}
    \caption{Comparison between different architectures designed for video tasks. (a) is the regular masked autoencoder \cite{tong2022videomae}, (b) is the DWT-based architecture with IDWT module, and (c) is our proposed method.}
    \label{fig:compare}
\end{figure}

Since DWT has been widely applied in noisy data reconstruction, achieving notable gains in both noise removal and complex tasks like deblurring \cite{williams2018wavelet,li2020wavelet}, our approach builds on the methodologies of WaveCNets \cite{Li_2020_CVPR} by incorporating DWT for noise-robust representation. Figure \ref{fig:compare}(b) shows IDWT configuration which computes attention using only low-frequency components of the query and key for noise filtering(similar to \cite{wu2024waveformer}) during attention computation. Similar to WaveCNets, we discard high-frequency components during the initial downsampling layer but extend this technique to 3D-DWT to handle temporal decomposition. Our framework leverages 3D-DWT’s ability to decompose both spatially and temporally, enabling effective noise handling across frames and enhancing robustness over the entire video sequence. As shown in Figure 1, our approach improves both accuracy and relative robustness compared to several action recognition models on the UCF-101 dataset.

\begin{figure*}[htbp]
    \centering
    \includegraphics[width=0.9\linewidth]{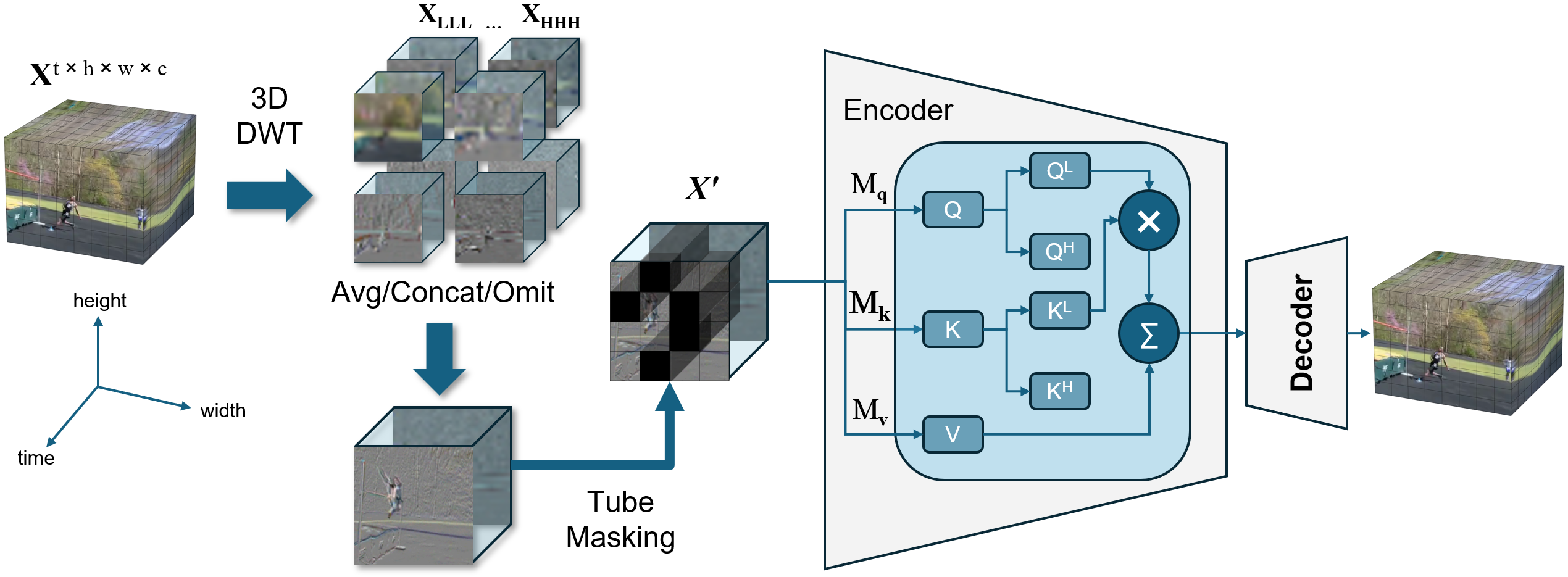}
    \caption{The framework of RobustFormer. Our approach integrates spatio-temporal tube masking as well as multi-resolution feature transformation using DWT to handle real-world noise types.}
    \label{fig:frame}
\end{figure*}

\section{Methods}
Compared to the classical VideoMAE \cite{tong2022videomae}, our proposed method enhances robustness by employing 3D-DWT for spatio-temporal analysis; in contrast to other IDWT-based methods, our approach eliminates the computationally expensive IDWT step, significantly improving efficiency, as illustrated in Figure \ref{fig:compare}(c). 

Our implementation incorporates a two-step wavelet transform. First, we utilize DWT’s efficient downsampling capability to generate embeddings for video patches. By leveraging 3D-DWT, we can process video data more effectively than stacking coefficients from 2D-DWT. This approach enables simultaneous handling of both temporal and spatial noise. The robust feature extraction achieved is thus the result of noise-adaptive compression in the initial DWT phase, noise filtering during attention computation well as masked pretraining which allows to recover back to RGB without requiring additional IDWT step. The forward and backward propagation processes for this step are detailed below. For experiments with images we use 2D-DWT instead on 3-dimensional version while every other aspects remaining same.

\subsection{Wavelet Embedding}
Let $\mathbf{X}^{t\times h\times w\times c}$ be the original downsampled and clipped video sequence of sequence length $t$, height $h$, width $w$ and channels $c$. We start with the generation of transformation matrices based on the selected wavelet filters. This step is critical as it defines the low-pass and high-pass filters that are applied in the transformation process. We define the wavelet by low-pass (L) and high-pass (H) filter coefficients as $\mathbf{L}=[l_1,l_2,\ldots,l_n],\, $ and $\mathbf{H}=[h_1,h_2,\ldots,h_n]$. In this work, we employ Haar wavelets due to their computational efficiency and effectiveness in noise reduction.

Now we construct the transformation matrices for each dimension (depth, height, and width). For simplicity, only the depth (time) dimension's matrix construction is shown here. We can replicate this for the height and width dimensions:
\begin{equation}
\small
\begin{aligned}
    &\mathcal{L}_k=\text{ConstructMatrix}(\mathbf{L},k)\\
    &\mathcal{H}_k=\text{ConstructMatrix}(\mathbf{H},k)\\
\end{aligned}    
\end{equation}
where, k $\in$ \{H, W, D\} represent each of the individual dimension. The forward pass thus involves applying these matrices to decompose the input data into its respective frequency sub-bands. The input tensor $\mathbf{X}$ is thus decomposed into eight sub-bands using the transformation matrices for depth $(\mathcal{L}_D,\mathcal{H}_D)$, height $(\mathcal{L}_H,\mathcal{H}_H)$, and width $(\mathcal{L}_W,\mathcal{H}_W)$, as shown in Figure \ref{fig:frame}.
\begin{equation}
    \small
    \begin{aligned}
        &\mathbf{X}_{LLL}=\mathcal{L}_D^T\mathcal{L}_H^T\mathcal{L}_W^T\mathbf{X},\\&\mathbf{X}_{LLH}=\mathcal{L}_D^T\mathcal{L}_H^T\mathcal{H}_W^T\mathbf{X},\\&\ldots,\\&\mathbf{X}_{HHH}=\mathcal{H}_D^T\mathcal{H}_H^T\mathcal{H}_W^T\mathbf{X}
    \end{aligned}
    \label{eq:DWT3d}
\end{equation}
where, the matrix construction leads to generating matrix $\mathcal{M}_k \in \{\mathcal{L}_k, \mathcal{H}_k\}$ for dimension corresponding to $k$. The resulting matrix $\mathcal{M}_k$ will have dimensions $\left(\left\lceil\frac k2\right\rceil,k\right)$
,where $\left\lceil\cdot\right\rceil$ is the ceiling function that handles cases where $k$ is odd. for filter $\mathbf{F} \in \{\mathbf{L}, \mathbf{H}\}$ and $\mathbf{F}=[f_1,f_2,\ldots,f_n]$. The ConstructMatrix function works as follows:

\begin{equation}
    \small
    \mathcal{M}^{ij}_k=\begin{cases}f_{t+1}&\mathrm{if} j=2i+t  \mathrm{,} t<n\\0&\mathrm{otherwise}\end{cases}
    \label{eq:DWT1d}
\end{equation}
where, where $t$ indexes the filter coefficients $[f_1,f_2,\ldots,f_n]$.


\begin{algorithm}[ht]
    \small
    \caption{RobustFormer Pseudocode}
    \label{algo:robustformer}
    \DontPrintSemicolon
\SetKwFunction{SelfAttention}{self attention}
\SetKwFunction{PatchEmbedDWT}{PatchEmbedDWT}
\SetKwProg{Fn}{Function}{:}{}
\SetKwInOut{Input}{Input}
\SetKwInOut{Output}{Output}


\BlankLine
\Fn{\PatchEmbedDWT{$x$}}{
    \For{$i\gets1$ \KwTo $\text{dims}$}{
        $dwt[i] \gets \text{DWT\_2D}(x[i])$\;
    }
    $embeddings\gets \text{Masking}(\text{Conv}(\text{wavelet\_ops}(dwt)))$\;

    \Return $embeddings$\;
}

\BlankLine
\textbf{Main:}\\
$embeddings \gets PatchEmbedDWT(x)$\;
$Z \gets \text{Encoder}(embeddings)$\;
\BlankLine
\textbf{Pre-training:}\\
$\hat{x} \gets \text{Decoder}(Z)$\;
$\mathcal{L}_{\text{recon}} \gets \text{MSE}(x,\,\hat{x})$\;
\BlankLine
\textbf{Fine-tuning:}\\
$\hat{y} \gets \text{RobostFormer}(Z)$\;
$\mathcal{L}_{\text{task}} \gets \text{CrossEntropy}(\hat{y},\,y)$\;
\BlankLine
Update model parameters via backpropagation\;

\end{algorithm}


\subsection{Attention Computation}

In the wavelet embedding step, we average, concatenate, or omit high-frequency components to reduce noise and smooth the input representation. However, the subsequent latent transformation (3D convolution with tube masking) may amplify residual noise during attention calculation. To minimize redundant details, we aim to reduce noise as much as possible during training. Prior work \cite{wu2024waveformer} has shown that noise before the attention mechanism distorts correlation scores by increasing values for unrelated pairs (due to random alignment) and decreasing values for related pairs (as noise weakens true alignment), ultimately impairing classification accuracy. To mitigate this, we follow \cite{wu2024waveformer} and fully omit high-frequency components, ensuring a smoother latent representation before attention calculation.

Suppose \(\boldsymbol{X}' = \{\boldsymbol{x}'_1, \boldsymbol{x}'_2, \ldots, \boldsymbol{x}'_T\}\) denotes the output of the tube-masked 3D convolution, representing deep encoded feature representations. Each \(\boldsymbol{x}'_i\) is a tensor in \(\mathbb{R}^{\frac{t}{t'} \times e \times \frac{h}{p} \times \frac{w}{p}}\), where \(t'\) denotes the tubelet size, \(e\) the embedding dimension, and \(p\) the patch size. The attention mechanism elements \(\mathbf{Q}\), \(\mathbf{K}\), and \(\mathbf{V}\) are computed by passing \(\boldsymbol{X}'\) through three distinct linear layers.

We then perform 1D-DWT for each $\textbf{Q}$, $\textbf{K}$ and $\textbf{V}$ by obtaining the low and high pass filters corresponding to a specific wavelet(Haar) similar to \cite{wu2024waveformer} as follows,
\begin{equation}
\small
    \begin{aligned}
        &\mathbf{Q}_{L}=\mathcal{L}_Q^T\mathbf{Q}, \mathbf{Q}_{H}=\mathcal{H}_Q^T\mathbf{Q}\\
        &\mathbf{K}_{L}=\mathcal{L}_K^T\mathbf{K}, \mathbf{K}_{H}=\mathcal{H}_K^T\mathbf{K}\\
    \end{aligned}
\end{equation}
We calculate these segregated components similar to eq (\ref{eq:DWT3d}) and omit $\mathbf{Q_H}$ and $\mathbf{K_H}$. Note that we also omit DWT computation for $\mathbf{V}$, so that the attention computation can attend to some useful high level components from $\mathbf{V}$ (that are omitted in $\mathbf{Q}$ and $\mathbf{K})$ in the input, as shown in Figure \ref{fig:frame}. With this we can avoid performing additional IDWT step as in \cite{wu2024waveformer} which makes the architecture complicated and computationally intensive. 

We finally compute our attention score as follows:
\begin{equation}
    \small
    \text{Attention}=\text{softmax}(\frac{\mathbf{Q_L}\mathbf{K_L^T}}{\sqrt{d_k}})\mathbf{V}
    \label{eq:attention}
\end{equation}
The backward function is defined for each DWT computation similar to 3D DWT computation.

With this we create various configurations of RobustFormer that involves various operations over DWT blocks. \textbf{RF-A} indicates RobustFormer with DWT averaging, \textbf{RF-AA} refers to averaging as well as DWT attention to every attention layer as mentioned in \ref{eq:attention}. Similarly we define \textbf{RF-O}, \textbf{RF-OA}, \textbf{RF-C} and \textbf{RF-CA} for omit,  omit using DWT attention, concat and concat using DWT attention.

\begin{table}[ht]
\small
\centering
\caption{Comparison of robustness of various Models using mean corruption error (mCE), normalized by AlexNet values, on ImageNet-C. `*' represents models with wavelet based strategy. }
\label{table1}
\begin{tabular}{lc}
\toprule
\textbf{Models} & \textbf{IN-C (mCE)} $\downarrow$ \\
\midrule
\textbf{CNNs} & \\
ResNet-50 \cite{he2016deep}  & 76.7 \\
ResNeXt50-32x4d \cite{xie2017aggregated} & 64.7 \\
VGG-11 \cite{simonyan2014very}  & 93.5 \\
VGG-19 \cite{simonyan2014very}  & 88.9 \\
VGG-19 + BN \cite{simonyan2014very}  & 81.6 \\
ANT \cite{rusak2020simple} & 63.0 \\
EWS \cite{guo2022improving}  & 63.0 \\
WResNet-18* \cite{li2020wavelet} & 80.8 \\
WResNet-101* \cite{li2020wavelet} & 65.8 \\
\midrule
\textbf{ViTs} & \\
PVT-Large \cite{wang2021pyramid} & 59.8 \\
BiT-mr101x3 \cite{kolesnikov2020big}  & 58.3 \\
MAE-ViT-B \cite{NEURIPS2022_416f9cb3} & 58.8 \\
\midrule
\textbf{Robust Formers (Ours)} &  \\
RobustFormer-A  & 55.3 \\
RobustFormer-AA & 55.7 \\
RobustFormer-O  & \textbf{55.3} \\
RobustFormer-OA & 55.9 \\
RobustFormer-C  & 57.7 \\
RobustFormer-CA & 58.3 \\
\bottomrule
\end{tabular}
\end{table}

\begin{table}[h!]
\small
\centering
\caption{Comparison of RobustFormer models with varying configurations using absolute mean Flip Probability(\textbf{mFP}) metric on Imagenet-P. `$\downarrow$' indicates lower is better.}
\label{tab:robustformer_comparison}
\begin{tabular}{lcccc}
\hline
\textbf{Model} & \textbf{Params} & \textbf{Flops} & \textbf{IN-P(mFP)} $\downarrow$ \\
\hline
 MAE-ViT-B \cite{NEURIPS2022_416f9cb3} &111.7M &36.7G & 13.4\\
RobustFormer-A   & 111.2M & 36.6G   & 12.7\\
RobustFormer-AA  & 111.2M & 36.6G   & 12.7\\
RobustFormer-O   & 111.6M & 37.5G   & 12.2\\
RobustFormer-OA  & 111.6M & 37.5G   & 12.4\\
RobustFormer-C   & 111.2M & 38.0G   & 12.9\\
RobustFormer-CA  & 111.2M & 38.0G   & 12.9\\
\hline
\end{tabular}
\end{table}

\begin{table*}[ht]
\centering
\small
\caption{Comparison of robustness of various networks under every type of distortion in imagenet-C using the corruption error metric(CE) \cite{hendrycks2018benchmarking} using AlexNet as the baseline. `*' represents models with Wavelet based strategy. The best models are marked as \textbf{BOLD} for each corruption category. Lower values are better.}
\label{table3}

\setlength{\tabcolsep}{1pt}

\begin{tabular}{lccccccccccccccc}
\toprule
\multirow{2}{*}{Network} & \multicolumn{3}{c}{Noise} & \multicolumn{4}{c}{Blur} & \multicolumn{4}{c}{Weather} & \multicolumn{4}{c}{Digital} \\

\cmidrule(lr){2-4} \cmidrule(lr){5-8} \cmidrule(lr){9-12} \cmidrule(lr){13-16}
  &  Gauss. & Shot & Impulse & Defocus & Glass & Motion & Zoom & Snow & Frost & Fog & Bright & Contrast & Elastic & Pixel & JPEG \\
\midrule
AlexNet      & 100 & 100 & 100 & 100 & 100 & 100 & 100 & 100 & 100 & 100 & 100 & 100 & 100 & 100 & 100  \\
SqueezeNet    & 107 & 106 & 105 & 100 & 103 & 101 & 100 & 101 & 103 & 97 & 97 & 98 & 106 & 109 & 134 \\
VGG-11        & 97  & 97  & 100 & 92  & 99  & 93  & 91  & 92  & 91  & 84  & 75 & 86 & 97 & 107 & 100  \\
VGG-19         & 89  & 91 & 95  & 89  & 98  & 90  & 90  & 89  & 86  & 75  & 68 & 80 & 97 & 102 & 94 \\
VGG-19+BN      & 82  & 83  & 88  & 82  & 94  & 84  & 86  & 80 & 78  & 69  & 61 & 74 & 94 & 85 & 83  \\
ResNet-18      & 87  & 88  & 91  & 84  & 91  & 87  & 89 & 86  & 84 & 78  & 69 & 78 & 90 & 80 & 85 \\
ResNet-50     & 80  & 82  & 83  & 75  & 89  & 78  & 80  & 78  & 75  & 66 & 57 & 71 & 85 & 77 & 77  \\

WResNet-18*   &80.2 & 80.5& 80.5& 79.7& 89.8& 83.6& 84.8& 84.9& 80.8& 73.9&66.3 & 75.1& 88.2 &75.1& 88.6 \\
WResNet-101*  & 64.3 & 65.9 & 65.0 & 63.6 & 80.4 & 68.2 & 71.8 & 71.6 & 67.5 & 60.2 & 50.1 & 62.9 & 74.6 & 52.3 & 67.9 \\

AugMix & 67 & 66 & 68 & 64 & 79 & 59 & 64 & 69 & 68 & 65 & 54 & 57 & 74 & 60 & 66\\

MAE-ViT   & 56.7& 57.3 & 56.1& 61.9& 75.4& 58.1& 68.6& 57.7 & 63.4 & 53.7& 45.4& 45.0& 68.9 & 55.1& 58.7\\
\hline
RF-O  & 51.9&  52.3 & 52.0 & \textbf{58.0}& 71.7& 56.0& 65.2& 57.1 & 58.9 & 54.4& \textbf{44.2} & 45.4 & 63.4 & \textbf{45.6}& 56.5\\

RF-OA & \textbf{51.6} & \textbf{51.8} & \textbf{51.8} & 59.2 & 72.2 & \textbf{55.9} & 64.8 & 56.9 & 60.2 & 55.8 & 44.4 & 45.3 & 64.2 & 47.9 & 56.4\\ 

RF-A & 52.9 & 53.5 & 52.4 & 58.7 & \textbf{68.7} & \textbf{55.9} & 64.4 & 56.6 & \textbf{58.6} & 55.3 & 44.5 & \textbf{43.7} & \textbf{61.6} & 47.8 & \textbf{55.2}\\

RF-AA & 52.8 & 53.9 & 52.8 & 58.9 & 69.1 & 57.4 & \textbf{64.1} & 56.8 & 59.5 &55.5 & 44.9 & 44.8 & 61.9 & 48.0 & 55.9\\ 

RF-C & 53.6 & 54.6 & 52.1 & 62.9 & 76.6 & 57.4 & 69.2 & \textbf{55.9} & 61.9 & \textbf{52.8} & \textbf{44.2} & 45.4 & 69.1 & 52.2 & 57.3 \\

RF-CA & 54.6 & 55.6 & 53.4 & 62.8 & 75.9 & 57.2 & 69.2 & 57.4 & 62.6 & 53.3 & 44.4 & 47.1 & 69.4 & 53.5 & 57.5 \\
\hline

\end{tabular}
\end{table*}

\begin{table*}[h!]
    \centering
    \small
    \caption{Comparison of robustness of Video Classification models on 5 different corruption categories: Noise, Blur, Temporal, Digital and Camera\cite{schiappa2023large} evaluated on Kinetic-400P benchmark dataset. $\gamma^a$ and $\gamma^r$ are absolute and relative robustness scores of the models averaged accross all the severity level and noise types in a particular category. The best models are marked as \textbf{BOLD} for each corruption category.  For both, higher is better.}
    \label{table4}
    \begin{tabular}{lcccccccccccccc}
        \toprule
        \multirow{2}{*}{\textbf{Network}} & \multirow{2}{*}{\textbf{Params}}& \multirow{2}{*}{\textbf{Flops}} & \multicolumn{2}{c}{\textbf{Noise}} & \multicolumn{2}{c}{\textbf{Blur}} & \multicolumn{2}{c}{\textbf{Temporal}} & \multicolumn{2}{c}{\textbf{Digital}} & \multicolumn{2}{c}{\textbf{Camera}} & \multicolumn{2}{c} {\textbf{Mean}} \\
        &&& \textbf{$\gamma^a$} & \textbf{$\gamma^r$} & \textbf{$\gamma^a$} & \textbf{$\gamma^r$} & \textbf{$\gamma^a$} & \textbf{$\gamma^r$} & \textbf{$\gamma^a$} & \textbf{$\gamma^r$} & \textbf{$\gamma^a$} & \textbf{$\gamma^r$} & \textbf{$\gamma^a$} & \textbf{$\gamma^r$} \\
        \midrule
        R3D\cite{hara2017learning} & 32.5M& 55.1G&.71 & .61 & .78 & .70 & .98 & \textbf{.97} & .91 & .88 & .89 & .85 & .85 & .80 \\
        I3D\cite{carreira2017quo} & 28.0M& 75.1G&.72 & .61 & .80 & .72 & .97 & .96 & .91 & .87 & .89 & .85 & .86 & .80 \\
        SF\cite{feichtenhofer2019slowfast}  & 34.6M& 66.6G&.73 & .64 & .81 & .73 & .96 & .94 & .91 & .87 & .88 & .84 & .85 & .80 \\
        X3D\cite{feichtenhofer2020x3d} & 3.8M& 5.15G&.71 & .62 & .81 & .75 & .96 & .94 & .90 & .86 & .85 & .84 & .85 & .80 \\
        TF\cite{bertasius2021space}  & 122M& 196G&.87 & .84 & \textbf{.91} & \textbf{.87} & \textbf{.98} & .96 & .94 & .94 & .95 & \textbf{.93} & .91 & .88 \\
        MViT\cite{fan2021multiscale} & 36.6M& 70.7G&.93 & .91 & .86 & .82 & .96 & .95 & .94 & .93 & .92 & .92 & .93 & .91 \\
        VideoMAE\cite{NEURIPS2022_416f9cb3} &94.8M&167.7G&\textbf{.95}&\textbf{.92}& .89& .79& .97& .95& .97& .95& .84& .69& .92&.86  \\
        \hline
        RF-A &93.2M&162.8G&.94&.90&.90&.82&.97&.95&.97&.94&.85&.69 & .93&.86  \\
        RF-AA &93.2M&162.8G&.94&.91&.90&.80&.96&.95&\textbf{.97}&.95&.85&.70&.92&.86 \\
        RF-O  &93.2M&160.4G&.94&.91&.90&.81&.97&\textbf{.97}&.96&\textbf{.96}&\textbf{.96}&.68 &\textbf{.95}&\textbf{.87}\\
        RF-OA &93.2M&160.4G&\textbf{.95}&.91&.90&.80&.97&.96&.97&.94.&.86&.70&.93&.86 \\

        \bottomrule
    \end{tabular}
\end{table*}


\begin{figure*}[ht]
    \centering
    \begin{subfigure}[b]{0.22\linewidth}
        \centering
        \includegraphics[width=\linewidth, trim=40 0 0 0, clip]{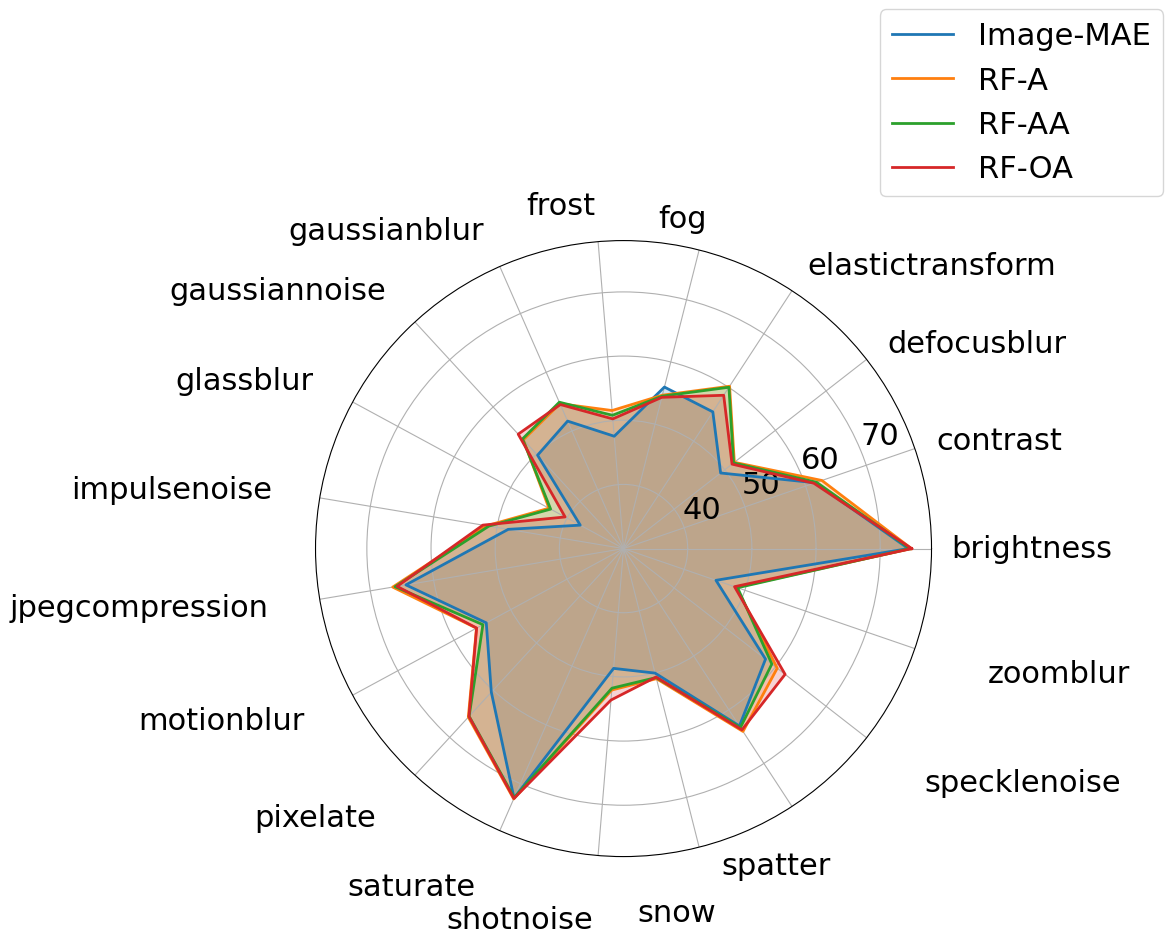}
        \caption{Top-1 Accuracy Imagent-C}
        \label{fig:top1_accuracy}
    \end{subfigure}
    \begin{subfigure}[b]{0.22\linewidth}
        \centering
        \includegraphics[width=\linewidth, trim=40 0 0 0, clip]{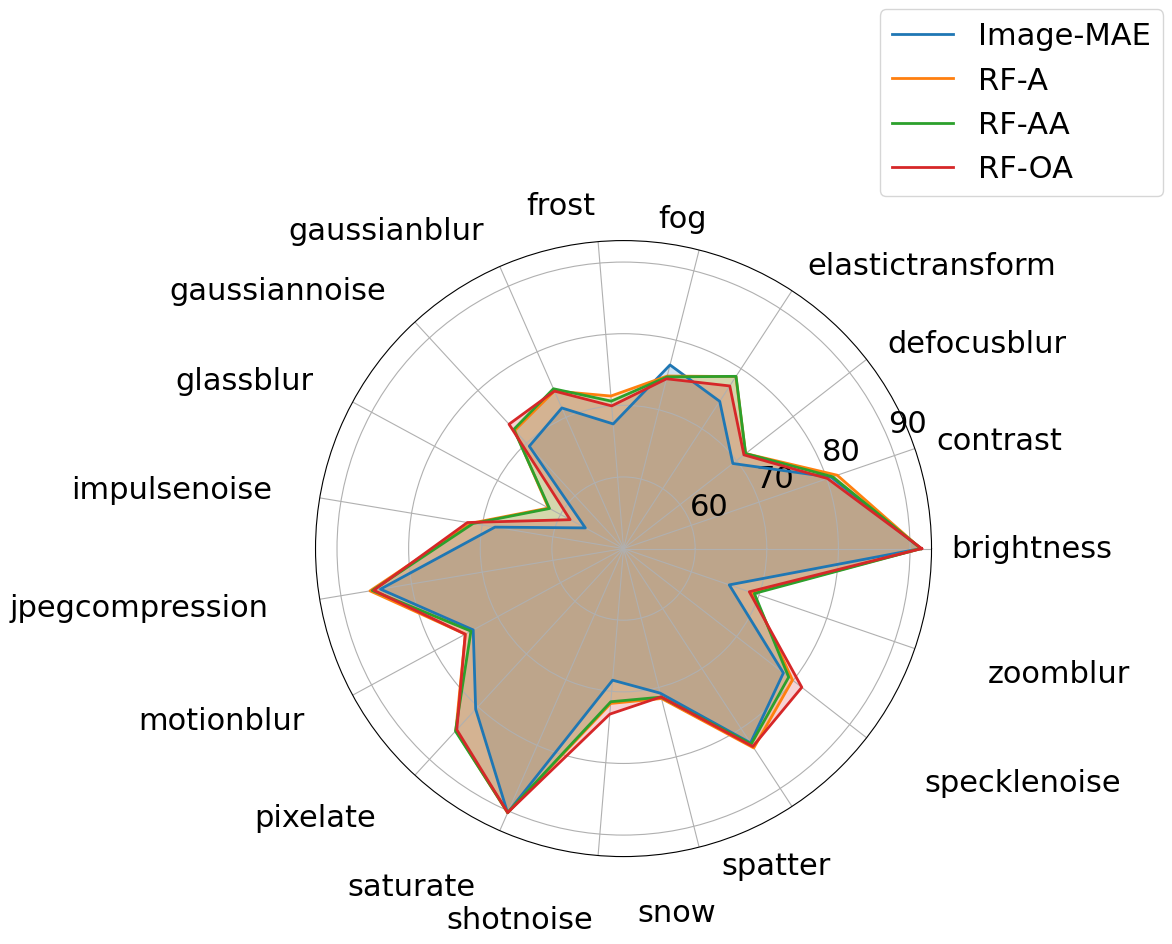}
        \caption{Top-5 Accuracy Imagenet-C}
        \label{fig:top1_accuracy}
    \end{subfigure}
    \begin{subfigure}[b]{0.25\linewidth}
        \centering
        \includegraphics[width=\linewidth, trim=20 40 150 0, clip]{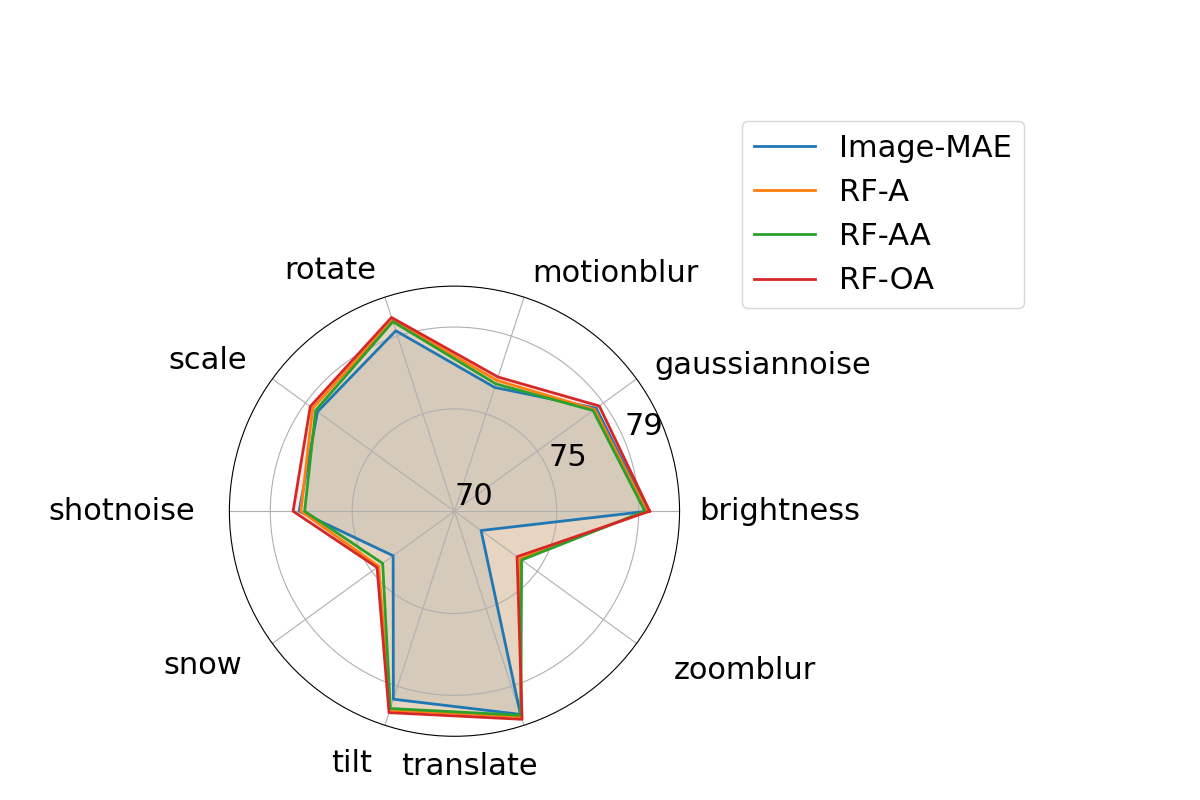}
        \caption{Top-1 Accuracy Imagenet-P}
        \label{fig:top1_accuracy}
    \end{subfigure}
    \begin{subfigure}[b]{0.25\linewidth}
        \centering
        \includegraphics[width=\linewidth, trim=20 40 150 0, clip]{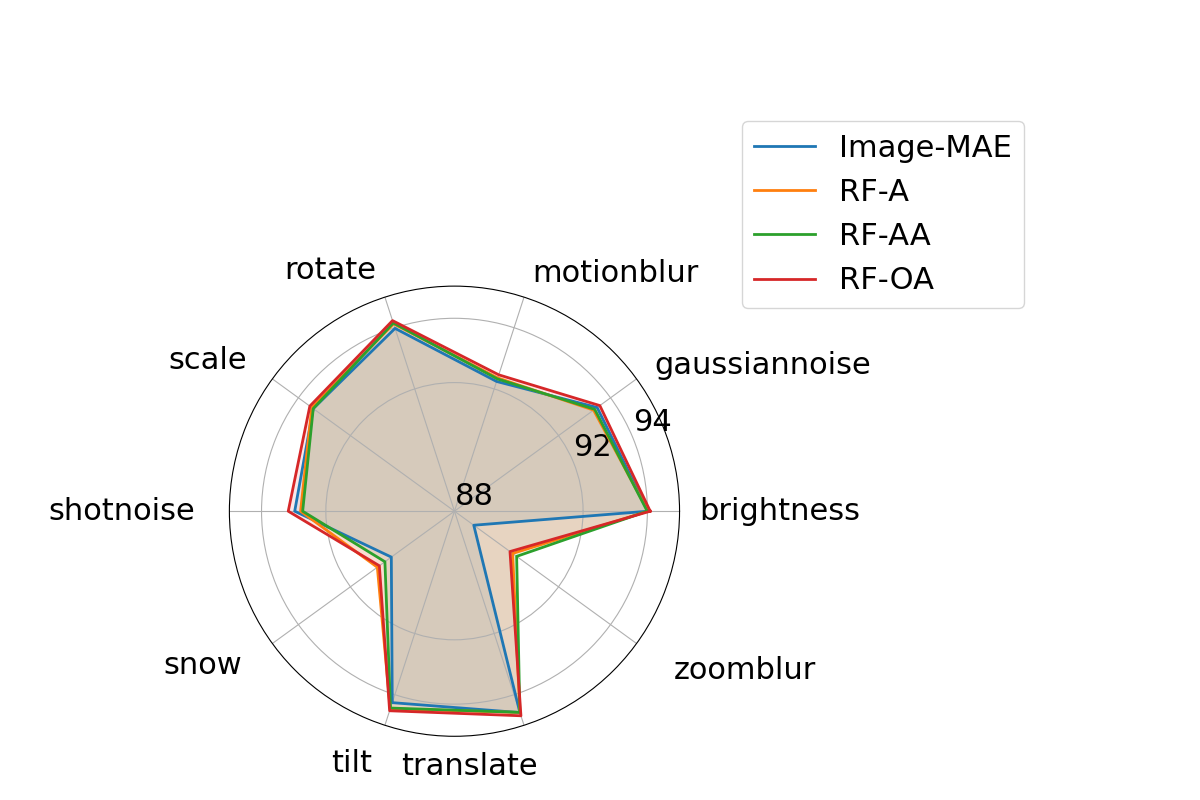}
        \caption{Top-5 Accuracy Imagenet-P}
        \label{fig:top1_accuracy}
    \end{subfigure}
    \caption{Comparison of Top-1 and Top-5 Accuracy for Imagenet-C and Imagenet-P. Accuracy for Imagent-C are averaged across severity levels}
    \label{fig:accuracy_comparison-imagenet-cp}
\end{figure*}

\begin{table*}[h!]
\centering
\resizebox{0.9\textwidth}{!}{ 

\begin{minipage}{0.7\textwidth}
    \small
    \caption{Relative robustness for individual corruptions ($\gamma^{r}_{p}$) averaged across all severity levels for Kinetics-400 dataset. The best models are marked as \textbf{BOLD} for each perturbation.}
    \label{table5}
    \setlength{\tabcolsep}{3pt}
    \begin{tabular}{lccccccc|ccc}
        \hline
        \textbf{Perturbation} & \textbf{R3D} & \textbf{I3D} & \textbf{SF} & \textbf{X3D} & \textbf{TF} & \textbf{MViT} & \textbf{VideoMAE} & \textbf{RF-A} & \textbf{RF-AA} & \textbf{RF-OA} \\
        \hline
        Defocus Blur  & .77 & .75  & .80 & .85 & .80 & .83 & .87&\textbf{.88}&.87&\textbf{.88}       \\
        Motion Blur  & .63& .60 & .64  & .66 & .75  & .82  & .85&\textbf{ .89}&\textbf{.89}&.88       \\
        Zoom Blur    & .71 & .74& .80 & .85  & .89  & .90  & .65& .65&.65&.65       \\
        Gaussian              & .47 & .46   & .36   & .49  & .75 & .87   & .88& .85&.85&.87       \\
        Shot                  & .78 & .79   & .82   & .79  & .94 & .95   &.96& .96&\textbf{.98}&.96      \\
        Impulse               & .44 & .42   & .34   & .46  & .76 & .87   &.87& .83&.84&.86     \\
        Speckle               & .75 & .75   & .70   & .75  & .91 & .95   &.95&.95&.95&\textbf{.96}      \\
        Compression           & .93 & .90   & .92   & .89  & .94 & .94   & .95&.95&.95&\textbf{.96}       \\
        Static Rotate         & .67 & .65   & .70   & .71  & .82 & .87   & .85&.86&.86&\textbf{.87}       \\
        Rotate                & .92 & .93   & .83   & .87  & \textbf{.97} & .90 & .77& .80&.83&.82       \\
        Translate             & .97 & .96   & .89   & .93  & \textbf{.99} & .95 &.43& .41&.41&.42       \\
        Jumbling              & \textbf{.97} & .96   & .89   & .91  & .95 & .91 &.89&.89&.92&.90       \\
        Box Jumbling          & \textbf{.99} & .97   & .96   & .96  & .97 & .94 &\textbf{.99}&\textbf{.99}&\textbf{.99}&\textbf{.99}       \\
        \hline
    \end{tabular}
\end{minipage}%
\hfill
\begin{minipage}{0.25\textwidth}
    \centering
    \footnotesize
    \caption{Parameters and FLOPs for Video models.}
    \label{tab:flops}
    \begin{tabular}{lc}
        \toprule
        \textbf{Model}  & \textbf{Flops} \\
        \midrule
        VideoMAE\cite{tong2022videomae} & 167.7G \\
        \hline
        RF-A & 162.8G \\
        RF-AA & 162.8G \\
        \textbf{RF-O}  & 160.4G \\
        \textbf{RF-OA  }& 160.4G \\
        \hline
        RF-I & 166G \\
        RF-IA & 167.5G \\
        \bottomrule
    \end{tabular}
\end{minipage}
}
\end{table*}

\section{Experiments}

    \subsection{Datasets}
We evaluate our method across five datasets: \textbf{Kinetics-400} \cite{kay2017kinetics}, \textbf{UCF-101} \cite{soomro2012ucf101} and \textbf{HMDB-51} for video tasks, and \textbf{ImageNet-1K} \cite{deng2009imagenet} and \textbf{ImageNet-Tiny-200} \cite{le2015tiny}, for image tasks. The \textbf{Kinetics-400} dataset includes 240k training videos and 20k validation videos, each clip lasting 10 seconds. The dataset focuses on human-centered actions, covering a wide range of interactions. The \textbf{UCF-101} dataset, while smaller with 13.3k action videos across 101 categories, offers high diversity in actions and significant variations in camera motion, object appearance, pose, background, and lighting conditions. For image-based experiments, we use the standard \textbf{ImageNet-1K} dataset.

For noisy evaluation on video datasets, we assess our models under real-world perturbations following \cite{schiappa2023large}, along with additional noise types and intensity configurations. Video noise types include four categories: \textbf{Noise} (Gaussian, shot, impulse, speckle), \textbf{Blur} (zoom, motion, defocus), \textbf{Digital} (JPEG and MPEG compression artifacts), \textbf{Temporal} (jumble, box jumble), and \textbf{Camera Motion} (static rotation, dynamic rotation, translation). Each noise type is tested across five severity levels, resulting in a total of \textbf{70} unique noise configurations. In the image experiments, we evaluate model robustness using the ImageNet-P and ImageNet-C benchmarks \cite{hendrycks2019benchmarking}. ImageNet-C includes 15 types of corruptions with five severity levels, totaling \textbf{75} distinct corruptions. ImageNet-P adds 10 perturbation types with 30 intensity variations, leading to \textbf{300} different noise scenarios. All pre-training, fine-tuning, and validation are conducted on clean datasets, without noise augmentation.

    \subsection{Training}

We use ViT-Base (ViT-B/16, input size 224) \cite{dosovitskiy2020image} as the backbone for all training and ablation studies, covering both image and video datasets. ViT-B, while over three times larger than ResNet-50 \cite{he2016deep}, is lightweight compared to other transformer architectures, making it an efficient yet powerful choice for our experiments. For image tasks, pretraining is conducted on ImageNet using a masking ratio of 0.75 and norm-pix-loss as the reconstruction objective, following the MAE framework \cite{he2022masked}. Training spans 400 epochs with a batch size of 64, distributed across 20 nodes equipped with P100 GPUs (2 GPUs and 28 cores per node). Fine-tuning is performed for 100 additional epochs in full float-32 precision without autocast.

\begin{figure*}[ht]
    \centering
    \includegraphics[width=\linewidth]{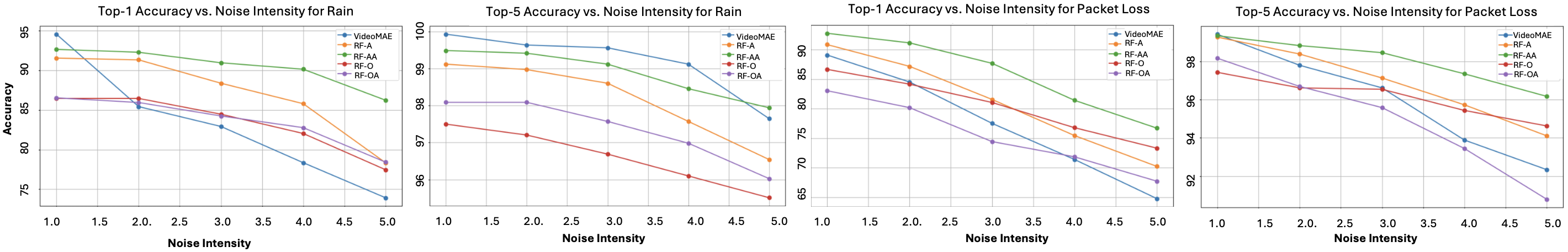}
    \caption{Comparison of Top-1 and Top-5 Accuracy for different severity levels across rain and packet loss noise for UCF-101 dataset}
    \label{fig:accuracy_comparison1}
\end{figure*}

\begin{figure*}[t]
    \centering
    \captionsetup[subfigure]{justification=centering} 

    \begin{subfigure}[t]{0.24\linewidth}
        \centering
        \includegraphics[height=3.2cm,keepaspectratio,trim=30 0 40 0,clip]{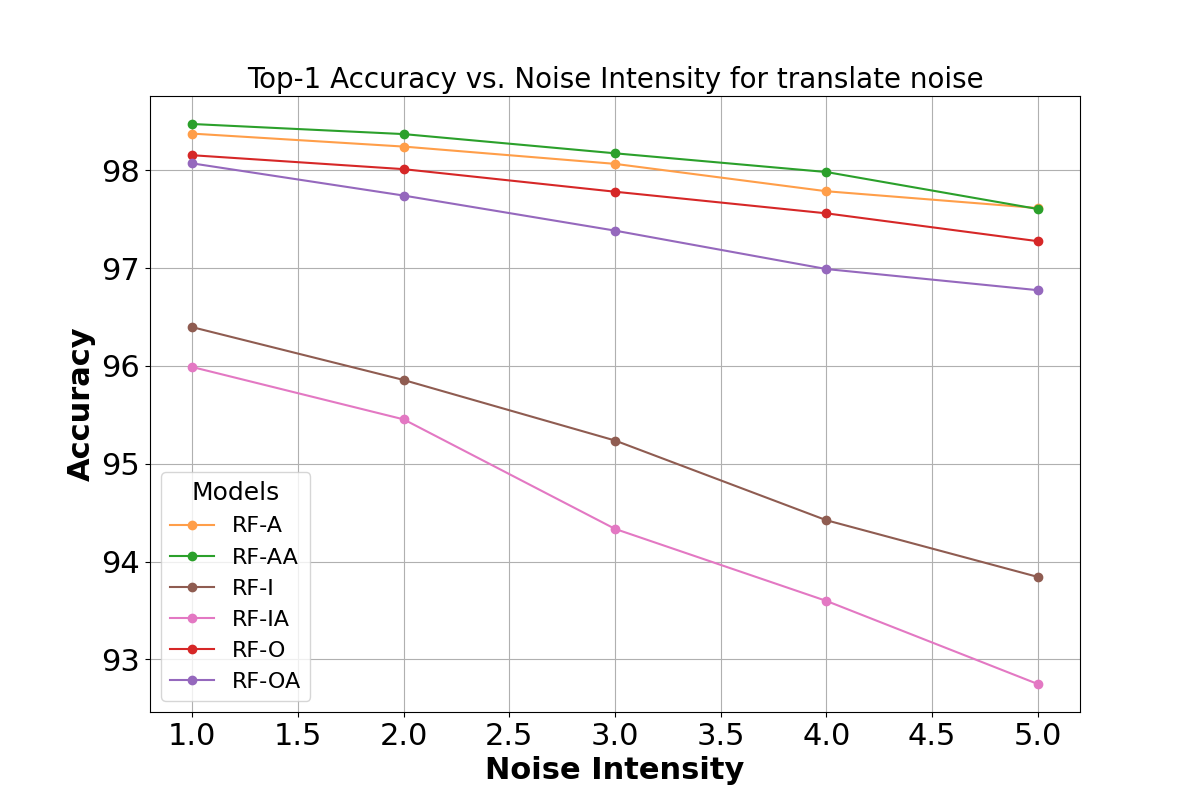}
        \caption{Top-1 accuracy for Translation noise}
        \label{fig:ablation_translation}
    \end{subfigure}\hfill
    \begin{subfigure}[t]{0.24\linewidth}
        \centering
        \includegraphics[height=3.2cm,keepaspectratio,trim=40 0 40 0,clip]{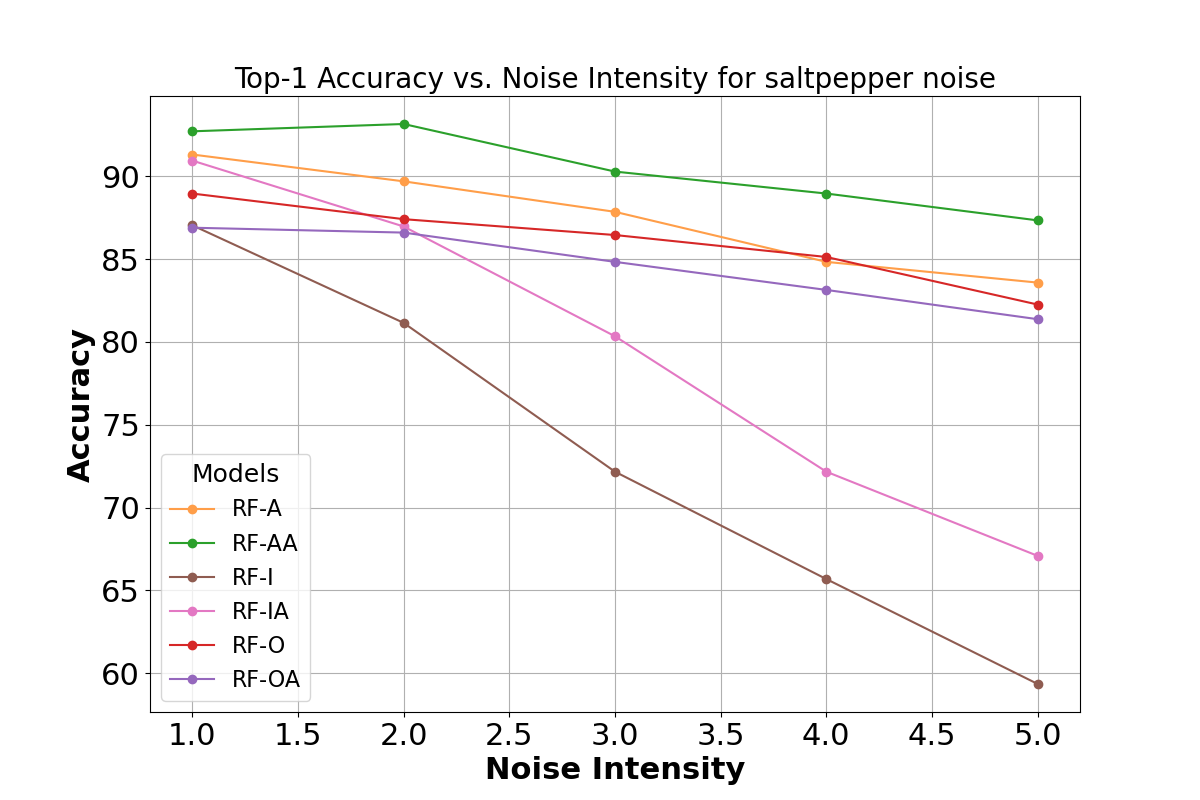}
        \caption{Top-1 accuracy for Salt-and-pepper noise}
        \label{fig:ablation_saltpepper}
    \end{subfigure}\hfill
    \begin{subfigure}[t]{0.24\linewidth}
        \centering
        \includegraphics[height=3.2cm,keepaspectratio,trim=40 0 40 0,clip]{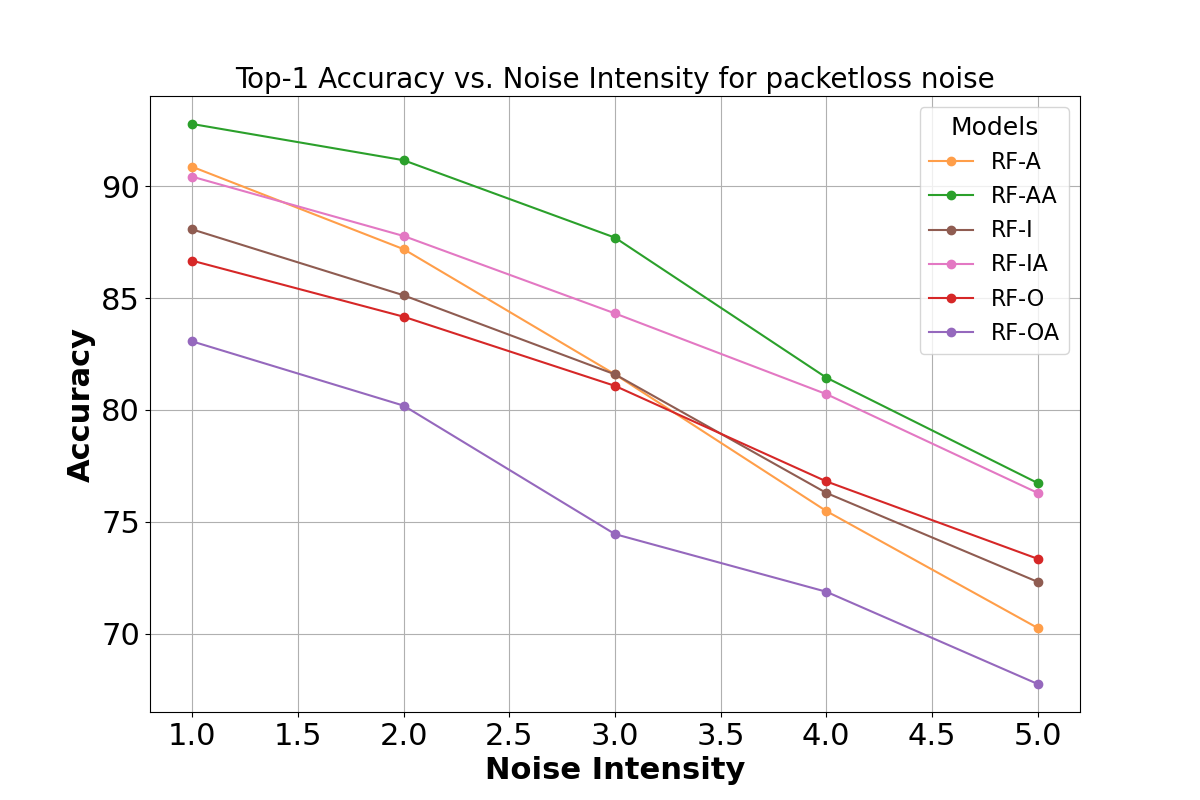}
        \caption{Top-1 accuracy for Packet loss}
        \label{fig:ablation_packetloss}
    \end{subfigure}\hfill
    \begin{subfigure}[t]{0.24\linewidth}
        \centering
        \includegraphics[height=3.2cm,keepaspectratio,trim=40 0 40 0,clip]{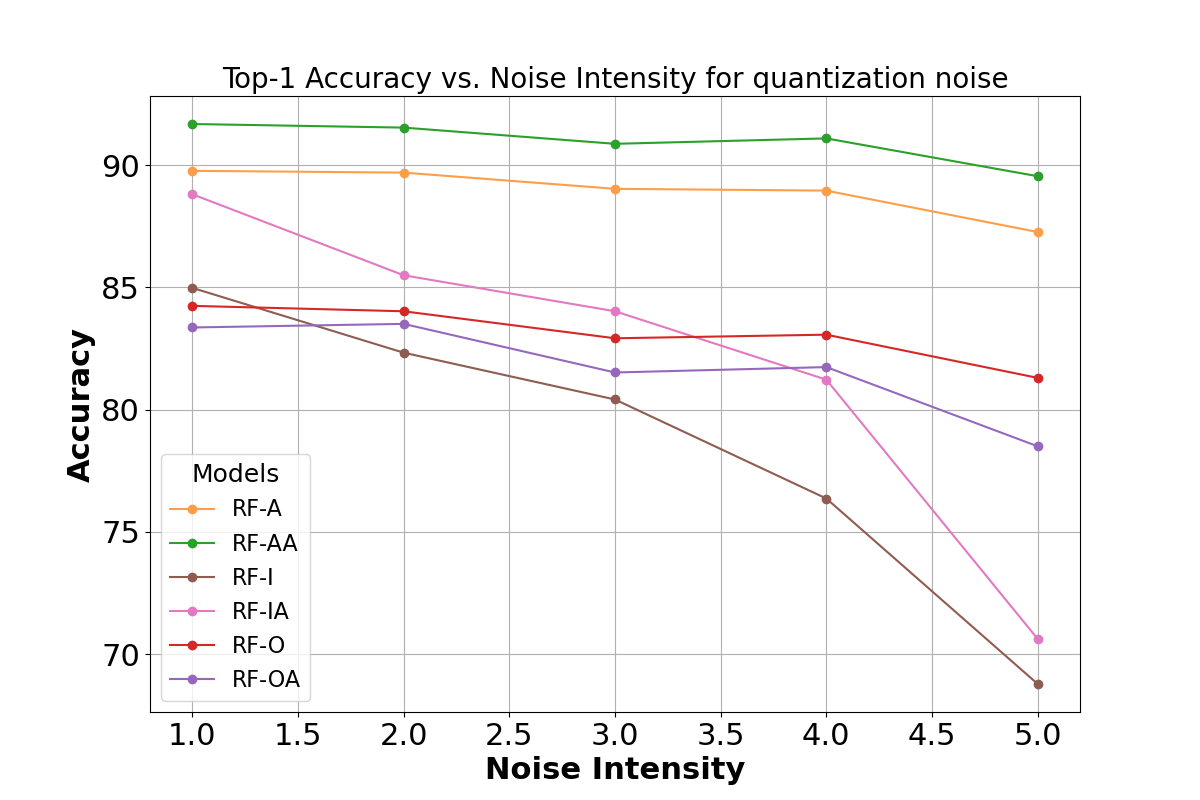}
        \caption{Top-1 accuracy for Quantization noise}
        \label{fig:ablation_quantization}
    \end{subfigure}

    \caption{Ablation experiments: Comparison of Top-1 accuracy across various noise types for variants with average, IDWT, and DWT attention.}
    \label{fig:ablation}
\end{figure*}

For video datasets, we apply the same foundational framework with adaptations tailored to temporal data. Pretraining on UCF-101 and HMDB-51 involves 800 epochs using 16-frame sequences, a batch size of 8, a tube masking ratio of 0.9, a sampling rate of 4, and a decoder depth of 4. The same node configuration is used as in the image tasks. Fine-tuning for these datasets is conducted over 100 epochs. For larger-scale datasets like Kinetics-400P, training is distributed across 12 nodes with V100 GPUs (2 GPUs and 40 cores per node) for 400 epochs, maintaining a tube masking ratio of 0.9, sampling rate of 4, and decoder depth of 4 but with an increased batch size of 16. We use DeepSpeed \cite{rasley2020deepspeed} during video fine-tuning for memory-saving optimizations.\\ 
Our evaluation focuses on the model's robustness to real-world noise, including shot noise, rain noise, Gaussian noise, packet loss, speckle noise, and tampering noise, as identified in prior benchmarks \cite{momeny2021noise, yi2021benchmarking}. These pertubations mirror real-world conditions, where noise is often diverse and unstructured. For instance, practical noise types like rain, critical for autonomous driving, and packet loss, which may affecting video-based applications. Additionally, testing on image datasets shows that our DWT-based approach is not only robust but also achieves a 2.5\% accuracy improvement on clean datasets.

    \subsection{Evaluation}
        For evaluating noise robustness on Imagenet-C we compute the mean corruption error(mCE) as follows:
        \begin{equation}
            \small
            \text{CE}_c^f = \left( \sum_{s=1}^{5} E_{s,c}^f \right) \Big/ \left( \sum_{s=1}^{5} E_{s,c}^{\text{Baseline}} \right)
        \end{equation}
        where, $E_{s, c}^f$ is the top-1 error for corruption ${c}$ and severity ${s}$ for the model ${f}$ and $E_{s, c}^{\text{Baseline}}$ is the same for the baseline model for which we use AlexNet. The mean of the corruption error(mCE) across corruptions of RobutFormer and other comparable models is shown in Table \ref{table1}, whereas, CE for the individual corruptions is shown in Table \ref{table3}. Variants of RobustFormer methods always beats MAE \cite{he2022masked} and CNN based approaches. Similarly, for Imagenet-P we compute the flip probability of the model `${f}$' on `${m}$' perturbation sequences ${S}$ as 
        \begin{equation}
            \small
            \mathrm{FP}_p^f = \frac{1}{m(n-1)} \sum_{i=1}^m \sum_{j=2}^n \mathbbm{1} \left( f\left(x_j^{(i)}\right) \neq f\left(x_1^{(i)}\right) \right)
        \end{equation}
        where $x_1^{(i)}$ is the clean image and $x_j^{(i)}$ with $(j>1)$ are the pertubed images of $x_1^{(i)}$. We then average across all perturbations to get mean flip error(mFP). The comparison of RobustFormer variants with Image-MAE\cite{he2022masked} is shown in Table \ref{tab:robustformer_comparison} where all the RF perform better. Figure \ref{fig:accuracy_comparison-imagenet-cp}, shows both the average top-1 and top-5 accuracies on Imagenet-P and Imagenet-C noise benchmarks which demonstrates that our method is robust to almost all evaluated noise types for both benchmarks.

        To measure video robustness we use two metrics for relative and absolute accuracy drops. After training model ${f}$, we first compute the accuracy $A_c^{f}$ on the clean set and accuracy $A_{p,s}^f$ for perturbation ${p}$ and severity ${s}$. The absolute robustness and relative robustness are then computed as $\gamma_{p,s}^a=1-(A_c^f-A_{p,s}^f)/100$ and $\gamma_{p,s}^r=1-(A_c^f-A_{p,s}^f)/A_c^f$. The aggregated performance for all models can be thus achieved by averaging across severity levels to get $\gamma_{p}^a$ and $\gamma_{p}^r$. Table \ref{table4} shows $\gamma^{a}$ and $\gamma^{r}$, which are averaged $\gamma_{p}^a$ and $\gamma_{p}^r$ across the perturbation categories $p$ for Kinetics-400 dataset. Our RF variants even without augmented training demonstrated significant robustness compared to some of the methods that used augmentation during training. Similarly, Table \ref{table5} shows the relative robustness scores $\gamma_{p}^{r}$ for individual perturbations averaged across 5 severity  levels. RF models here demonstrated better relative robustness to recent state-of-the-art methods. Moreover, top-1 and top-5 accuracy performance on additional noise types like rain, packetloss for UCF-101 dataset are shown in Figure \ref{fig:accuracy_comparison1}, where RF methods are significantly superior especially in noise with high severity levels. Additional results (including for ImageNet-Tiny-200 dataset evaluated on our custom noise) are kept in supplementary.

\subsection{Ablation Study}
We ablate the effects of the IDWT step and the proposed DWT-based attention module across different RobustFormer variants. Our preferred model, \textbf{RF-AA}, achieves competitive and often superior top-1 and top-5 accuracy compared to all other configurations. The variants \textbf{RF-I} and \textbf{RF-IA} correspond to adding the IDWT reconstruction step and adding both IDWT and DWT-attention, respectively. While the IDWT step is commonly used in traditional DWT pipelines, it offers only limited gains when used alone.

Our experiments show that the \textbf{DWT-attention module is the primary source of robustness improvements}. Averaging alone struggles with high-impulse corruptions such as salt-and-pepper noise, whereas adding DWT attention yields substantial accuracy gains. Moreover, DWT attention continues to improve performance even within IDWT-based variants (Fig.~\ref{fig:ablation}b--d). Removing the IDWT step also reduces computational cost by \textbf{7.1 GFLOPs} when comparing RF-OA with RF-IA (Table~\ref{tab:flops}). Overall, these ablations highlight the central role of DWT-based attention in achieving strong robustness under diverse noise conditions.

\section{Discussion and Conclusion}
Noise-robust models are essential for many real-world applications, where noise patterns can significantly impact performance. While some deep learning models rely on heavy augmentation and others on rule-based filtering, these approaches are limited in scalability due to the inherent randomness and diversity of noise. Our approach enables efficient decomposition of spatio-temporal information, selectively focusing on low-frequency components to enhance robustness across diverse noise types without the need for exhaustive augmentation or complex filtering rules, thus making few assumptions about the evaluation noises. We validated our method on both image and video benchmark datasets with benchmark noise types for both and show our model's significance compared to prior works.

{\small
\bibliographystyle{ieee_fullname}
\bibliography{references}
}

\end{document}